\documentclass[10pt,twocolumn,letterpaper]{article}

\usepackage[pagenumbers]{iccv} %

\usepackage[accsupp]{axessibility}  %
\usepackage{bm}
\usepackage{subcaption}
\usepackage{ifthen}
\usepackage{pifont}
\usepackage{siunitx}

\usepackage{float}
\usepackage{tabularx}
\newcolumntype{Y}{>{\centering\arraybackslash}X}

\usepackage{pgfplots}
\pgfplotsset{compat=1.18}

\usepackage{tikz}
\usetikzlibrary{fit,positioning,arrows.meta,decorations.markings,calc,fit,decorations.pathreplacing,shadows,plotmarks}
\pgfdeclarelayer{background}
\pgfsetlayers{background,main}

\usepackage{xcolor}
\definecolor{RWTHSchwarz50}{RGB}{156,158,159}
\definecolor{RWTHBlau100}{RGB}{0,84,159}
\definecolor{RWTHMagenta100}{RGB}{227,0,102}
\definecolor{RWTHGelb100}{RGB}{255,237,0}
\definecolor{RWTHGruen100}{RGB}{87,171,39}
\definecolor{RWTHOrange100}{RGB}{246,168,0}
\definecolor{RWTHRot100}{RGB}{204,7,30}
\definecolor{RWTHViolett100}{RGB}{97,33,88}

\definecolor{QualitativeRed}{RGB}{178,34,34}
\definecolor{QualitativeGreen}{RGB}{44,160,44}

\newcommand{\down}{$\downarrow$}

\newcommand{\cmark}{\ding{51}}%
\newcommand{\xmark}{\ding{55}}%

\newcolumntype{s}{>{\centering}p}

\definecolor{TableGrey}{RGB}{150,150,150}
\newcommand{\grey}[1]{\textcolor{TableGrey}{#1}}

\newcommand{\paragraphsmallerspace}[1]{
    \vspace{-0.3cm}
    \paragraph{#1}
}

\usepackage{scalerel}
\def\DONUT{\scalerel*{\includegraphics{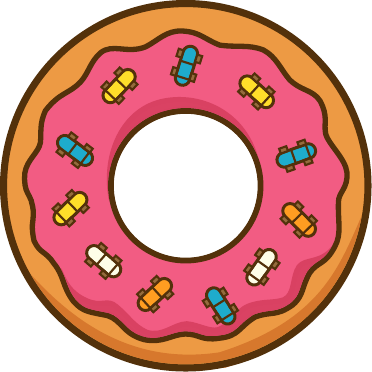}}{\textrm{O}}}

\newcommand{\PAR}[1]{\vskip4pt \noindent {\bf #1~}}

\expandafter\def\expandafter\normalsize\expandafter{%
    \normalsize%
    \setlength\abovedisplayskip{4pt}%
    \setlength\belowdisplayskip{4pt}%
    \setlength\abovedisplayshortskip{4pt}%
    \setlength\belowdisplayshortskip{4pt}%
}

\newcommand{\httpsurl}[1]{\href{https://#1}{\texttt{#1}}}

\definecolor{iccvblue}{rgb}{0.21,0.49,0.74}
\usepackage[pagebackref,breaklinks,colorlinks,allcolors=iccvblue]{hyperref}

\def\paperID{8865} %
\def\confName{ICCV}
\def\confYear{2025}

\title{D\DONUT{}NUT: A Decoder-Only Model for Trajectory Prediction}

\author{
Markus Knoche$^1$ %
\quad
Daan de Geus$^{1,2}$%
\quad
Bastian Leibe$^1$\\[5pt]
$^1\hspace{1pt}$RWTH Aachen University \quad $^2\hspace{1pt}$Eindhoven University of Technology\\[5pt]
\httpsurl{vision.rwth-aachen.de/donut}
}

\begin{document}
\maketitle
\begin{abstract}

Predicting the motion of other agents in a scene is highly relevant for autonomous driving, as it allows a self-driving car to anticipate. Inspired by the success of decoder-only models for language modeling, we propose DONUT, a Decoder-Only Network for Unrolling Trajectories. Unlike existing encoder-decoder forecasting models, we encode historical trajectories and predict future trajectories with a single autoregressive model. This allows the model to make iterative predictions in a consistent manner, and ensures that the model is always provided with up-to-date information, thereby enhancing performance. Furthermore, inspired by multi-token prediction for language modeling, we introduce an `overprediction' strategy that gives the model the auxiliary task of predicting trajectories at longer temporal horizons. This allows the model to better anticipate the future and further improves performance. Through experiments, we demonstrate that our decoder-only approach outperforms the encoder-decoder baseline, and achieves new state-of-the-art results on the Argoverse 2 single-agent motion forecasting benchmark.

\end{abstract}

\section{Introduction}

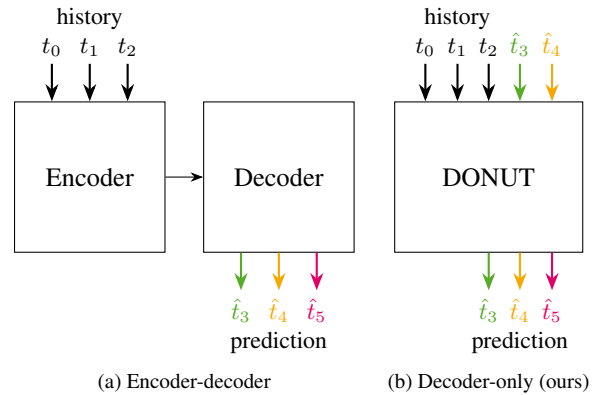
\begin{figure}
    \centering
    \tikzset{
    flow/.style={
        -{Stealth}
    },
    traj/.style={
        -{Stealth},
        thick
    },
}

\begin{subfigure}{0.55\linewidth}
    \centering
    \begin{tikzpicture}
        \node[draw, minimum height=2cm, minimum width=2cm] (encoder) {Encoder};
        \node[draw, minimum height=2cm, minimum width=2cm, right=0.5 of encoder] (decoder) {Decoder};
        \draw[traj] ($(encoder.north)!0.5!(encoder.north west)$) ++(0,0.5) -- ($(encoder.north)!0.5!(encoder.north west)$) node[near start,above,font=\small,yshift=0.1cm]{$t_0$};
        \draw[traj] (encoder.north) ++(0,0.5) -- (encoder.north) node[near start,above,font=\small,yshift=0.1cm]{$t_1$};
        \draw[traj] ($(encoder.north)!0.5!(encoder.north east)$) ++(0,0.5) -- ($(encoder.north)!0.5!(encoder.north east)$) node[near start,above,font=\small,yshift=0.1cm]{$t_2$};
        \draw[flow] (encoder) -- (decoder);
        \draw[traj,RWTHGruen100] ($(decoder.south)!0.5!(decoder.south west)$) -- ++(0,-0.5) node[near end,below,font=\small,yshift=-0.1cm]{$\hat t_3$};
        \draw[traj,RWTHOrange100] (decoder.south) -- ++(0,-0.5) node[near end,below,font=\small,yshift=-0.1cm]{$\hat t_4$};
        \draw[traj,RWTHMagenta100] ($(decoder.south)!0.5!(decoder.south east)$) -- ++(0,-0.5) node[near end,below,font=\small,yshift=-0.1cm]{$\hat t_5$};

        \node[font=\small] at ($(encoder.north)+(0,1.1)$) {history};
        \node[font=\small] at ($(decoder.south)+(0,-1.2)$) {prediction};
    \end{tikzpicture}
    \caption{Encoder-decoder}
    \label{fig:teasered}
\end{subfigure}
\begin{subfigure}{0.4\linewidth}
    \centering
    \begin{tikzpicture}
        \node[draw, minimum height=2cm, minimum width=2.5cm] (arnet) {DONUT};
        \draw[traj] ($(arnet.north)!0.67!(arnet.north west)$) ++(0,0.5) -- ($(arnet.north)!0.67!(arnet.north west)$) node[near start,above,font=\small,yshift=0.1cm]{$t_0$};
        \draw[traj] ($(arnet.north)!0.33!(arnet.north west)$) ++(0,0.5) -- ($(arnet.north)!0.33!(arnet.north west)$) node[near start,above,font=\small,yshift=0.1cm]{$t_1$};
        \draw[traj] (arnet.north) ++(0,0.5) -- (arnet.north) node[near start,above,font=\small,yshift=0.1cm]{$t_2$};
        \draw[traj,RWTHGruen100] ($(arnet.north)!0.33!(arnet.north east)$) ++(0,0.5) -- ($(arnet.north)!0.33!(arnet.north east)$) node[near start,above,font=\small,yshift=0.1cm]{$\hat t_3$};
        \draw[traj,RWTHOrange100] ($(arnet.north)!0.67!(arnet.north east)$) ++(0,0.5) -- ($(arnet.north)!0.67!(arnet.north east)$) node[near start,above,font=\small,yshift=0.1cm]{$\hat t_4$};
        \draw[traj,RWTHGruen100] (arnet.south) -- ++(0,-0.5) node[near end,below,font=\small,yshift=-0.1cm]{$\hat t_3$};
        \draw[traj,RWTHOrange100] ($(arnet.south)!0.33!(arnet.south east)$) -- ++(0,-0.5) node[near end,below,font=\small,yshift=-0.1cm]{$\hat t_4$};
        \draw[traj,RWTHMagenta100] ($(arnet.south)!0.67!(arnet.south east)$) -- ++(0,-0.5) node[near end,below,font=\small,yshift=-0.1cm]{$\hat t_5$};
        
        \node[font=\small] at ($(arnet.north)!0.33!(arnet.north west)+(0,1.1)$) {history};
        \node[font=\small] at ($(arnet.south)!0.33!(arnet.south east)+(0,-1.2)$) {prediction};
    \end{tikzpicture}
    \caption{Decoder-only (ours)}
    \label{fig:teaserdo}
\end{subfigure}
    \caption{\textbf{Encoder-decoder vs.~decoder-only methods for motion forecasting.} In contrast to existing works, which use an encoder-decoder architecture, DONUT uses a unified, autoregressive model to process agents' historical and future trajectories. This allows it to predict trajectories at different time steps in a consistent manner and receive up-to-date information of relevant scene elements, improving its performance.}
    \label{fig:teaser}
\end{figure}

Motion forecasting is a crucial task for autonomous driving. By predicting the future trajectories of agents in a scene, a self-driving car can anticipate and adjust its behavior. To forecast agents' motion accurately, a model needs to take into account the geometry of the surrounding road, the presence of other agents, and an agent's historical behavior. For this purpose, most recent motion-forecasting works~\cite{zhou2023qcnet,zhang2024demo,lan2024sept, ruan2024learning, tang2024hpnet} employ a model with two different modules (\cref{fig:teasered}): (1) an \textit{encoder} that embeds the historical trajectories into a latent space while considering the road layout and agent interactions, and (2) a \textit{decoder} that takes these historical embeddings and predicts the future trajectories, again taking into account road geometry and agent interactions. While these \textit{encoder-decoder} methods obtain impressive results, we posit that a \textit{decoder-only} model is better suited for this task, as it allows us to treat historical and future trajectories in a \textit{unified} manner. 

In existing encoder-decoder methods, agents' future trajectories are predicted by the decoder. Based on the historical embeddings generated by the encoder, the decoder typically predicts the future trajectory either (a) in its entirety, in a single pass~\cite{lan2024sept, tang2024hpnet, ruan2024learning}, or (b) recurrently~\cite{zhou2023qcnet, seff2023motionlm, rowe2023fjmp}. Both of these configurations have limitations. When predicting the entire trajectory in a single pass, the network is not adequately aware of the scene elements towards the end of the predicted trajectories. We expect this to negatively impact the accuracy of `far-future' predictions, leading to the use of anchors or goal points in some models to mitigate this~\cite{lin2024eda, wang2023prophnet, wang2023ganet}. This limitation is addressed in part by using a recurrent method~\cite{zhou2023qcnet}, where the decoder predicts future trajectories in an iterative manner, being updated about relevant scene points for each future sub-prediction. However, the feature representation of input data received by this decoder varies between different time steps, which unnecessarily complicates the decoder's task. Moreover, it receives only the original historical embeddings of other agents as provided by the encoder, which become outdated when predicting `far-future' trajectories, again limiting the ability to make an informed prediction. See \cref{sec:method:baseline} for more details.

In this work, we propose a method that overcomes these limitations by treating the historical time steps and future time steps in a unified manner, using a \textit{decoder-only network} (\cref{fig:teaserdo}). We call this approach the \textit{Decoder-Only Network for Unrolling Trajectories} (DONUT). In DONUT, we use a single network module to process both historical and future trajectories in an autoregressive manner. In each step, the model is informed of the scene elements---road geometry and other agents---around the endpoint of the previously predicted trajectory. As such, the network is aware of the relevant scene elements---including other agents' trajectories---at all times. Furthermore, because of the autoregressive decoder-only formulation, each future trajectory prediction is made based on the embeddings of the trajectories at all previous time steps, generated by the same model. As such, each prediction is made in a consistent manner. We expect that this allows the model to naturally transfer what it learns about historical trajectories to the future, and that it simplifies the task for the model, enabling it to predict future trajectories more accurately.

Our decoder-only approach to future trajectory prediction has similarities with autoregressive Large Language Models (LLMs)~\cite{radford2018gpt}. Interestingly, recent works~\cite{gloeckle2024multitoken} have demonstrated that an LLM's output quality can be improved by tasking the model with \textit{multi-token prediction}. During training, the LLM additionally predicts tokens that are further in the future, supervised with an auxiliary loss. This stimulates the model to better take into account possible futures during training. Inspired by this finding, we propose to adapt this concept for trajectory forecasting. Concretely, when predicting a future trajectory segment, we let the model not only predict the current segment, but also the next segment. We expect that this \textit{overprediction} strategy enables DONUT to look further into the future when making its predictions, allowing it to better anticipate and predict a more consistent future trajectory. Furthermore, we introduce a refinement module to enhance the initial predictions, which receives updated information about other scene elements at the endpoint of the initially predicted trajectory.

Through experimental evaluation, we find that our decoder-only model outperforms the encoder-decoder baseline, especially on the minFDE metric that considers the accuracy of the forecasted trajectory's endpoint. Moreover, we find that our overprediction strategy further improves performance. This demonstrates the value of our proposed approach. Compared to existing top-performing methods, DONUT also performs considerably better on the minFDE and brier-minFDE metrics and obtains new state-of-the-art results on the Argoverse 2 benchmark~\cite{wilson2021argoverse2}.

In summary, our contributions are as follows:
\begin{itemize}
    \item We present DONUT, an autoregressive decoder-only model for motion forecasting, which uses the same unified model to process historical and future trajectories.
    \item We propose the \textit{overprediction} strategy, where a model is asked to predict further into the future to better anticipate different future possibilities.
    \item With our approach, we achieve state-of-the-art performance on the Argoverse 2 benchmark.
\end{itemize}

\section{Related Work}

\paragraph{Motion forecasting.}
Early motion forecasting models \cite{lee2017desire, chai2020multipath, cui2019multimodal, hong2019rules, tang2019multiple, gilles2021home, phan2020covernet, salzmann2020trajectron++} rasterize maps and use convolutional neural networks to extract features about the surroundings. As this results in precision losses, modern approaches follow VectorNet \cite{gao2020vectornet} and LaneGCN \cite{liang2020lanegcn} and directly use vectorized data. Such models are commonly based on graph convolutions \cite{liang2020lanegcn, mohamed2020social-stgcnn, zeng2021lanercnn, gilles2022thomas, gao2023heterogcn} or use attention \cite{liu2021mmtransformer, yu2020star, zhou2023qcnet, shi2022mtr, wang2023prophnet, zhang2024demo, nayakanti2023wayformer, lin2024eda} for interaction between scene elements.

After all scene elements are encoded and have interacted with each other in an encoder, many older methods often either use recurrent models \cite{lee2017desire, hong2019rules, tang2019multiple, salzmann2020trajectron++} or shallow feedforward networks \cite{gao2020vectornet, liang2020lanegcn, nayakanti2023wayformer, zhou2022hivt} to predict the output trajectories. Recent architectures \cite{ngiam2022scenetransformer, gilles2022thomas, shi2022mtr, zhou2023qcnet, zhang2024demo} commonly use attention-based decoders to allow for further interaction between agents, between different modes, and with the road layout information. Some works \cite{zhou2023qcnet, zhou2024smartrefine, choi2023r-pred, tang2024hpnet} also incorporate an additional refinement module which enhances the model's initial predictions. In this work, we diverge from the common encoder-decoder structure and instead present a decoder-only model, called DONUT. We apply a refinement module after each prediction step to improve the trajectory and combine this with an overprediction strategy.

Many current approaches use an agent-centric coordinate system~\cite{tang2019multiple, zhou2022hivt, shi2022mtr, wang2023prophnet, zhao2021tnt}. A downside of this approach is that all scene elements have to be converted into each agent's reference frame separately, which comes at an extra cost. An alternative is the use of a single global coordinate system for all agents, which is employed for most raster-based methods but also some more recent models~\cite{ngiam2022scenetransformer}. However, this approach is not position-invariant, meaning that the prediction will change if the whole scene is shifted or rotated. Recently, QCNet~\cite{zhou2023qcnet} proposed a query-centric approach, where each scene element---\ie, road lane or trajectory segment---has its own reference frame, and relative positional encodings are used for interaction between scene elements. We also utilize the query-centric approach in DONUT, as this allows to re-use features from previous time steps without the need to re-compute them if any position information gets updated.

There are only a few approaches that do not have a distinct agent encoder, like DONUT. Social LSTM \cite{alahi2016sociallstm}, one of the oldest deep-learning-based motion forecasting models, uses an LSTM \cite{hochreiter1997lstm} on the full trajectory of pedestrians in a traffic scene. STAR~\cite{yu2020star} uses an attention-based model which predicts a single time step using the full history. To predict longer sequences, the prediction is concatenated to the observed history and the full updated trajectory is fed through the model again.

In the related area of motion simulation \cite{montali2023simagents}, a few recent models \cite{philion2024trajeglish, wu2025smart, zhou2025behaviorgpt} use GPT-style decoder-only models. Different from our approach, these models always predict independent samples from the underlying future distribution. In motion forecasting, we are interested in a fixed number of predictions which should span the possible future as well as possible. As such, these tasks require different approaches. In addition, encoder-decoder approaches trained for motion forecasting can be applied autoregressively for simulation \cite{nayakanti2023wayformer, montali2023simagents, jiang2024scenediffuser}. This has some conceptual similarity to DONUT's overprediction strategy, as the training horizon extends further than the required horizon during inference. However, these approaches differ as DONUT makes short overpredictions for each individual future token during training, instead of overpredicting the entire future just once, in one pass.

\PAR{Relation to Large Language Models.}
The Transformer architecture \cite{vaswani2017transformer} was initially proposed in the context of language modeling. While this architecture contains an encoder and a decoder, encoder-only \cite{devlin2019bert} or decoder-only \cite{radford2018gpt} models were introduced shortly afterwards. Recently, most well-known large language models \cite{touvron2023llama2, team2023gemini, achiam2023gpt4, liu2024deepseek3} use the decoder-only approach, as the causal behavior is a good fit for sequence prediction. As motion forecasting is also a sequence prediction task, this inspired us to apply a decoder-only model. To guide LLMs to better anticipate the future, multi-token prediction~\cite{gloeckle2024multitoken} tasks a model with predicting tokens for future time steps. In DONUT, we adapt this to the trajectory forecasting task, with `overprediction'.

\section{Method}
\label{sec:method}

\begin{figure*}
    \centering
    \input{tikz/preamble}

\def\alldist{0.45}
\def\rerefdist{1.4}
\def\arrowlabelshift{0.08cm}

\begin{tikzpicture}
    \node[scene] (propin) {
        \begin{tikzpicture}[rotate=\scenerotation, anchor=center]
            \drawrefs{fill=RWTHGruen100}{2}{0}{0}
            \drawtrajs{traj}{1}{0}{0}{0}
        \end{tikzpicture}
    };
    \coordinate[above=0.6cm of propin.north] (top);
    \coordinate[below=1.3cm of propin.south] (bottom);
    
    \node[label] at ($(propin|-top)$) {Previous\\sub-trajectory};
    
    \node[net, right=\alldist of propin] (prop) {Proposer};
    \node[scene, right=\alldist of prop.south east] (propout) {
        \begin{tikzpicture}[rotate=\scenerotation, anchor=center]
            \drawtrajs{traj,RWTHSchwarz50}{3}{0}{1}{1}
            \drawrefs{fill=RWTHGruen100}{2}{0}{0}
            \drawtrajs{traj}{2}{0}{0}{1}
        \end{tikzpicture}
    };
    \node[label] (proptop) at ($(prop|-top)$) {Agent tokens \\ from previous steps};
    \node[label] (propbottom) at ($(prop|-bottom)$) {To next step};
    
    \node[scene, right=\rerefdist of propout] (refin) {
        \begin{tikzpicture}[rotate=\scenerotation, anchor=center]
            \drawtrajs{traj,opacity=0}{3}{0}{0}{1}
            \drawrefs{fill=RWTHGelb100}{3}{0}{1}
            \drawtrajs{traj}{2}{0}{0}{1}
        \end{tikzpicture}
    };
    \node[net, anchor=west] at ($(prop-|refin.east) + (\alldist, 0)$) (ref) {Refiner};
    \node[label] (reftop) at ($(ref|-top)$) {Agent tokens \\ from previous steps};
    \node[label] (refbottom) at ($(ref|-bottom)$) {To next step};
    
    \node[draw, circle, font=\tiny, inner sep=0pt, right=\alldist of ref] (add) {$+$};
    \node[scene, right=\alldist of add] (refout) {
        \begin{tikzpicture}[rotate=\scenerotation, anchor=center]
            \drawtrajs{traj,RWTHSchwarz50}{3}{0}{0}{1}
            \drawrefs{fill=RWTHGelb100}{3}{0}{1}
            \drawtrajs{traj}{2}{0}{0}{0}
        \end{tikzpicture}
    };
    \node[label] at ($(refout|-top)$) {Predicted \\ sub-trajectory};
    
    \node[scene, right=\rerefdist of refout] (next) {
        \begin{tikzpicture}[rotate=\scenerotation, anchor=center]
            \drawtrajs{traj,opacity=0}{3}{0}{0}{1}
            \drawrefs{fill=RWTHOrange100}{3}{0}{0}
            \drawtrajs{traj}{2}{0}{0}{0}
        \end{tikzpicture}
    };
    \node[label] (nextbottom) at ($(next|-bottom)$) {Input to next\\decoder step};

    \draw[flow] (propin) -- (prop);
    \draw[flow] (prop) -- (propout);
    \draw[reref] (propout) -- (refin) node[midway,above,font=\footnotesize,align=center,xshift=\arrowlabelshift] {Adjust \\ ref. point};
    \draw[flow] (refin) -- (ref);
    \draw[flow] (ref) -- (add);
    \draw[flow] (add) -- (refout);
    
    \draw[flow] (propout) -- ($(propout.south)+(0, -0.2)$) -| (add);
    
    \draw[reref] (refout) -- (next) node[midway,above,font=\footnotesize,align=center,xshift=\arrowlabelshift] {Adjust \\ ref. point};

    \draw[flow] (prop.south) -- (propbottom);
    \draw[flow] (ref.south) -- (refbottom);
    \draw[flow] (next.south) -- (nextbottom);
    \draw[flow] (proptop) -- (prop.north);
    \draw[flow] (reftop) -- (ref.north);
    
    \draw[flow] ($(prop.east)!0.8!(prop.north east)$) -- ($(ref.west)!0.8!(ref.north west)$);
\end{tikzpicture}
    \caption{\textbf{DONUT architecture overview.} 
    Previously predicted sub-trajectories are fed through a proposer module to make a proposal prediction (\protect\tikz \protect\draw[traj, thick] (0,0.1) to[out=0, in=160] (0.5,0);) and an overprediction (\protect\tikz \protect\draw[trajover, thick] (0,0.1) to[out=0, in=160] (0.5,0);). The reference point for all relative encodings is then moved to the endpoint of the proposed trajectory (from \protect\tikz \protect\node[fill=RWTHGruen100, circle, inner sep=0pt, minimum size=5pt] {}; to \protect\tikz \protect\node[fill=RWTHGelb100, circle, inner sep=0pt, minimum size=5pt] {};). Next, the refiner predicts offsets which are added to the proposed trajectories to obtain the final predicted sub-trajectory and overprediction. Before using the predicted sub-trajectory as input to the next decoder step, the reference point is updated to the refined endpoint again (from \protect\tikz \protect\node[fill=RWTHGelb100, circle, inner sep=0pt, minimum size=5pt] {}; to \protect\tikz \protect\node[fill=RWTHOrange100, circle, inner sep=0pt, minimum size=5pt] {};). For details on the proposer and refiner, see \cref{fig:architecture_proposer}.}
    \label{fig:architecture}
\end{figure*}

We present DONUT, a decoder-only Transformer model for motion forecasting.
In this section, we first provide a detailed task definition, then describe the encoder-decoder baseline that we improve upon, and finally present our new method. The overall architecture is visualized in \cref{fig:architecture}.

\subsection{Task definition}
\label{sec:method:task_def}

Motion forecasting aims to predict the future locations of agents based on their historic behavior and a given scene context. We define the agent's historic behavior as $\bm{X} \in \mathbb{R}^{N\times T_\text{hist}\times C}$, where $N$ represents the number of agents in the scene, $T_\text{hist}$ is the number of observed time steps, and $C$ is the size of the state of the agent. This state may describe an agents' position, velocity, motion vector, \etc. In addition, each agent's semantic type (\eg, car, pedestrian) is known. Given the historic behavior $\bm{X}$, the task objective is to predict each agent's position at $T_\text{fut}$ future time steps. To achieve this, a model is allowed to make $K$ different predictions which are commonly referred to as `modes'. In other words, the desired output is a set of $K$ future trajectories for all $N$ agents, given by $\bm{Y} \in \mathbb{R}^{N\times K \times T_\text{fut}\times C_\text{pos}}$, where $C_\text{pos} = 2$, as we require each agent's future $(x,y)$ positions. Additionally, a model should assign a probability to each of the $K$ predictions, so it should output $\bm P \in \mathbb R^{N\times K}$ with $\sum_k \bm P_{n, k}=1 \forall n$.

As additional scene context, the road geometry of the scene is available. Each road lane is encoded as a set of polylines, each consisting of a varying number of $(x,y)$ coordinates. Relations between road lanes (\eg, connecting or parallel lanes), and lanes' semantic categories (\eg, bus lane or part of an intersection) are also known.

\subsection{Encoder-decoder baseline}
\label{sec:method:baseline}

We build our method upon QCNet~\cite{zhou2023qcnet}, a top-performing encoder-decoder method for motion prediction. As such, this method serves as our baseline.

\PAR{Query-centric scene encoding.}
Both the baseline and our method use a query-centric scene encoding. In this framework, all scene elements---\ie, agents' trajectories or road lanes---are encoded as tokens that can be processed by a Transformer model~\cite{vaswani2017transformer}. Each token uses its own local reference coordinate system, which means the method is invariant to global positions and orientations. Likewise, instead of using absolute positional encodings, relations between scene elements are encoded in a relative manner. For this, distance, direction, and relative orientation are (1) encoded as Fourier features \cite{vaswani2017transformer, tancik2020fourier}, (2) fed into a multi-layer perceptron (MLP), and (3) added to the respective key and value vectors in the attention operation. If the communicating scene elements belong to different time steps, the time step difference is also encoded.

\PAR{Map encoder.}
In this baseline, each scene's road lanes are encoded into a set of tokens through a \textit{map encoder}. Each point on a polyline $m$ is represented by computing Fourier features of the relative distance to the subsequent point and feeding this into an MLP, and adding a learned embedding of the point's semantic category. Then, an embedding for the polyline's category attends to these point-level tokens using relative positional encodings, resulting in a single token per polyline that embeds information about all its constituent points. Finally, self-attention enables interaction between different polylines in a certain radius. An embedding encodes the type of the relation within the attention, \eg, whether a polyline is the successor of another or just closeby. This yields the final map tokens $\bm{T}_{\text{M}} \in \mathbb{R}^{M \times D}$, where $D$ is the embedding dimension, and $M$ is the number of polylines. In DONUT, we use the same map encoder.

\PAR{Agent encoder.} 
The baseline further employs an \textit{agent encoder} to represent agents' historical trajectories as tokens $\bm{T}_{\text{A}} \in \mathbb{R}^{N \times T_\text{hist} \times D}$. Here, for each time step $t_\text{hist} \in \{1, ..., T_\text{hist} \}$, the encoder obtains each motion vector and velocity at each time step, computes Fourier features, passes them through an MLP, and adds a categorical learned embedding to obtain initial tokens $\bm{T'}_{\text{A}}$. Next, each token first interacts with previous tokens of the same agent, then with road tokens $\bm{T}_{\text{M}}$, and finally with other agents' tokens of the same time step. All interactions are modeled as attention with relative positional encodings. This is repeated twice and yields the updated agent tokens $\bm{T}_{\text{A}}$.

\PAR{Recurrent decoder.}
To predict future trajectories, the encoder-decoder baseline takes the map tokens $\bm{T}_{\text{M}}$ and agent tokens $\bm{T}_{\text{A}}$, and feeds them through another module, the \textit{recurrent decoder}. In this decoder, learned mode embeddings $\bm{E} \in \mathbb{R}^{{N}\times{K}\times{D}}$ are introduced, which interact with the agent tokens and map tokens. Concretely, each embedding attends to (1) the historical tokens of the corresponding agent, (2) the map tokens, and (3) the most recent historical token of other agents in the scene. This is repeated $2\times$ and followed by a self-attention layer across the $K$ different modes. The output embeddings $\bm{E'}$ are used to predict the trajectory $\bm{\hat Y'}_\text{dec} \in \mathbb{R}^{N \times K \times T_\text{dec} \times C_\text{pos}}$ for the next $T_\text{dec} \leq T_\text{fut}$ time steps, for all agents and modes. 

This process is repeated until there is a prediction for all $T_\text{fut}$ time steps, $\bm{\hat Y'}_\text{fut} \in \mathbb{R}^{N \times K \times T_\text{fut} \times C_\text{pos}}$. In these subsequent iterations, the learned embeddings $\bm{E}$ are replaced by the decoder's output embeddings $\bm{E'}$ of the previous iteration. Note that this introduces inconsistencies, as the decoder either takes (a) a learned embedding or (b) an earlier output embedding as input, and has to learn to handle both cases. We hypothesize that this limits performance, as it complicates the decoder's task. Moreover, in later iterations, the historical agent tokens $\bm{T}_{\text{A}}$ are not updated with the previous predictions. As a result, the agent encoder only provides the decoder with `outdated' information when predicting further into the future. We expect that this harms the accuracy of the baseline's \textit{far-future} prediction.

Finally, QCNet employs a refinement module on the full proposed trajectory $\bm{\hat Y'}_\text{fut}$. This module tokenizes the initial predictions with a gated recurrent unit (GRU)~\cite{cho2014gru}, attends to the historic agent tokens $\bm{T}_{\text{A}}$ and map tokens $\bm{T}_{\text{M}}$, and outputs an offset to the original predictions. This results in refined trajectories $\bm{\hat{Y}}_\text{fut} \in \mathbb{R}^{N \times K \times T_\text{fut} \times C_\text{pos}}$. 

\begin{figure*}
    \centering
    \input{tikz/architecture_proposer.tex}
    \caption{\textbf{Proposer architecture.}
    The input sub-trajectory is first tokenized relative to the reference point (\protect\tikz \protect\node[fill=RWTHGruen100, circle, inner sep=0pt, minimum size=5pt] {};). Then, the tokens attend to (1) sub-trajectory tokens from previous decoder steps, (2) map tokens, (3) nearby agents, and (4) other modes of the same agent. All attention operations use relative positional encodings based on the current reference point (\protect\tikz \protect\node[fill=RWTHGruen100, circle, inner sep=0pt, minimum size=5pt] {};). Finally, a detokenizer outputs the next sub-trajectory and an overprediction. The refiner model has the exact same architecture, only the inputs and outputs differ.}
    \label{fig:architecture_proposer}
\end{figure*}

\subsection{DONUT}
\label{sec:method:ours}

Our decoder-only network, DONUT, uses the same query-centric formulation as the baseline, but it does not employ an encoder-decoder structure that requires different modules to embed agents' histories and predict future trajectories. Instead, we propose to use a single, unified decoder to process agents' trajectories in an autoregressive manner, for both historical and future time steps. This allows the decoder to use the same procedure to make future predictions at any time step, avoiding any inconsistencies. Moreover, it ensures that the decoder is always provided with `up-to-date' information about agents' trajectories, allowing it to make more informed predictions about the far future.

\PAR{Overall architecture.}
To process and predict trajectories in an autoregressive manner, we split all agents' trajectories into sub-trajectories of $T_{\text{sub}}$ time steps. These sub-trajectories are then fed through our decoder-only model. Starting with the predicted sub-trajectories of the previous time step, $\bm{\hat Y}_{\{-T_{\text{sub}};0\}} \in \mathbb{R}^{N\times K \times T_\text{sub}\times C_\text{ph}}$, our model uses a \textit{proposer} module to predict the sub-trajectories of the next $T_{\text{sub}}$ time steps, $\bm{\hat{Y'}}_{\{0;T_{\text{sub}}\}} \in \mathbb{R}^{N\times K \times T_\text{sub}\times C_\text{ph}}$. Furthermore, to enable the network to better anticipate possible futures, we also task it to predict the trajectory for the $T_{\text{sub}}$-long sub-trajectories that follow the initial predictions $\bm{\hat{Y'}}_{\{0;T_{\text{sub}}\}}$. This results in an \textit{overprediction} which we denote by $\bm{\hat{Y'}}^\text{over}_{\{T_{\text{sub}};2T_{\text{sub}}\}}$. These overpredictions are supervised during training, but discarded during inference. Next, we update the reference point to the endpoint of initial prediction $\bm{\hat{Y'}}_{\{0;T_{\text{sub}}\}}$ and recompute the relative positional encodings to other scene elements. These are used by the refinement module together with the tokenized initial predictions and features from the proposer to predict offsets to the initially predicted trajectories. Applying these offsets yields refined predictions $\bm{\hat{Y}}_{\{0;T_{\text{sub}}\}}$ and refined overpredictions $\bm{\hat{Y}}^\text{over}_{\{T_{\text{sub}};2T_{\text{sub}}\}}$.
This process is visualized in \cref{fig:architecture}.

The output of each decoder step is a set of predicted trajectories $\bm{\hat{Y}}_{\{0;T_{\text{sub}}\}}$ and overpredicted trajectories $\bm{\hat{Y}}^\text{over}_{\{T_{\text{sub}};2T_{\text{sub}}\}}$. The predicted trajectories are then stored and used as an input to the decoder in the next step, where the trajectory for the subsequent $T_{\text{sub}}$ time steps will be predicted. This is continued until all sub-trajectories have been processed and all future trajectories $\bm{\hat{Y}} \in \mathbb{R}^{N \times K \times T_\text{fut} \times C_\text{ph}}$ have been predicted. Besides the position, we also predict each agent's heading at each time step, so $C_\text{ph}=3$. Our model requires this information to appropriately update the reference coordinate frame.

\paragraphsmallerspace{Proposer and refiner.}
The proposer and refiner use the same architecture to make their predictions (see \cref{fig:architecture_proposer}). First, the input sub-trajectories are encoded as tokens by the \textit{tokenizer}. For each of the $T_\text{sub}$ time steps of a sub-trajectory, this tokenizer extracts the position, heading, motion vector, and velocity with respect to the reference coordinate frame. For this coordinate frame, we use the endpoint of the sub-trajectory. Following our baseline, we compute Fourier features for each time step separately. To merge the $T_\text{sub}$ time steps into a single token, we pass each time step's features through an MLP, concatenate the $T_\text{sub}$ resulting embeddings, and process them by another MLP, finally outputting a token for each sub-trajectory. This is combined with an embedding of the agent type. We denote these tokens as $\bm{T}_\text{sub} \in \mathbb{R}^{N\times K \times D}$.

In the remainder of the module, these tokens are subjected to various types of attention. \textit{(1) Temporal self-attention:} Each token $\bm{T}_\text{sub} \in \mathbb{R}^{N\times K \times D}$ attends to all historical sub-trajectory tokens of the same agent. These historical tokens are retrieved from the proposer and refiner modules from the previous decoder steps to keep them up-to-date, as visualized in \cref{fig:architecture_proposer}. 
\textit{(2) Map attention:} The tokens cross-attend to the tokens from road lanes within a radius $r$ of the reference point. To obtain these tokens, we use the same map encoder as the baseline. \textit{(3) Social attention:} To make agents aware of the behavior of other agents, each token self-attends to the tokens of all other agents within radius $r$ that have the same mode. \textit{(4) Mode attention:} Finally, to ensure that there is interaction between the $K$ different modes, tokens self-attend to tokens that have different modes but belong to the same agent at the same time step. This whole process is conducted twice in both proposer and refiner.

After this factorized attention, the tokens are decoded with a \textit{detokenizer} to predict the next sub-trajectories and generate the overpredictions using an MLP. An additional MLP outputs a logit for each mode of each agent at the final time step of the refiner, which we use to compute the mode probabilities.

To initialize the model with the historic tokens $\bm{X} \in \mathbb{R}^{N\times T_\text{hist}\times C}$, we feed historic trajectories into both proposer and refiner. These historic trajectories can be processed in parallel for all agents and time steps. As the only purpose is to produce historic agent tokens for the temporal attention, we can discard the predictions. We run this in a unimodal setting, \ie, $K=1$. After the last time step of the historic trajectory, we duplicate the tokens along the mode dimension. From this point on, the model is multimodal. To allow the models to distinguish different modes, we add a mode embedding for each mode within the mode attention. As the modes diverge over time, we also make the mode attention aware of the time step since starting the multimodal predictions, using a time embedding.

\paragraphsmallerspace{Training.}
During training, we supervise the predicted and overpredicted trajectories of the $T_\text{fut}$ future time steps. Following HiVT~\cite{zhou2022hivt} and QCNet~\cite{zhou2023qcnet}, the positions that are predicted as part of $\bm{\hat{Y}} \in \mathbb{R}^{N\times K \times T_\text{fut}\times C_{\text{ph}}}$ are parametrized using a mixture of Laplace distributions. For an agent $n \in \{1, ..., N\}$, for each time step $t \in \{1, ..., T_\text{fut}\}$, and for each mode $k \in \{1, ..., K\}$, the model outputs the Laplace parameters location $\bm \mu_{n,k,t}^\text{pos}$ and scale $\bm b_{n,k,t}^\text{pos}$, together with mode probabilities $\bm P_{n, k}$. This results in a likelihood
\begin{equation}
    \label{eq:laplace}
    p(\bm{\hat Y}_{n}^\text{pos})=\sum_{k=1}^K\bm P_{n, k}\prod_{t=1}^{T_\text{fut}}\text{Laplace}(\bm{\hat Y}_{n,t}^\text{pos}|\bm \mu_{n,k,t}^\text{pos}, \bm b_{n,k,t}^\text{pos}).
\end{equation}
Additionally, to update the reference point at each decoder step, DONUT requires the agent's heading. For this, we use the von-Mises distribution. Specifically, the model outputs location $\bm \mu_{n,k,t}^\text{hd}$ and concentration $\bm \kappa_{n,k,t}^\text{hd}$, and the likelihood is given by
\begin{equation}
    \label{eq:vonmises}
    p(\bm{\hat Y}_{n}^\text{hd})=\sum_{k=1}^K\bm P_{n, k}\prod_{t=1}^{T_\text{fut}}\text{vonMises}(\bm{\hat Y}_{n,t}^\text{hd}|\bm \mu_{n,k,t}^\text{hd}, \bm \kappa_{n,k,t}^\text{hd}).
\end{equation}
Following QCNet, we optimize the parameters of the Laplace and von-Mises distributions ($\bm \mu_{n,k,t}^\text{pos}$, $\bm b_{n,k,t}^\text{pos}$, $\bm \mu_{n,k,t}^\text{hd}$, $\bm \kappa_{n,k,t}^\text{hd}$) only for the mode where the proposed trajectory has the smallest endpoint distance to the ground truth. The loss is applied for proposed and refined trajectories independently, for both main predictions and overpredictions, and gradients are stopped after the output of the proposed trajectory. We apply the same procedure for the overpredictions. The mixture components $\bm P_{n,k}$ are optimized using the joint negative log-likelihood of \cref{eq:laplace} and \cref{eq:vonmises}. All losses are weighed equally.

\section{Experiments}
\label{sec:experiments}

\subsection{Experimental setup}
\label{sec:experiments:setup}

\paragraph{Datasets.} 
We assess the effectiveness of our model on the competitive Argoverse 2 single-agent forecasting benchmark~\cite{wilson2021argoverse2}. We train on the \textit{train} set, which consists of 200k scenes, and evaluate on the \textit{val} set and the hidden \textit{test} set, each containing 25k scenes. Each scene is \SI{11}{\second} long, sampled at \SI{10}{\hertz}. Of these \SI{11}{\second}, the model takes the trajectory of the first \SI{5}{\second} as input and predicts $K=6$ trajectories for the last \SI{6}{\second}. In other words, there are $T_\text{hist}=50$ historical time steps and $T_\text{fut}=60$ future time steps.

\PAR{Metrics.} 
We use the benchmark's default metrics for evaluation~\cite{wilson2021argoverse2}. The \textit{minimum final displacement error} ($\text{minFDE}_{K}$) computes the L2 distance between the endpoint of the ground-truth trajectory and the endpoint of the `best' predicted trajectory, across the $K$ predicted modes. Here, a predicted trajectory is the `best' if it has the lowest endpoint error.  Similarly, the \textit{minimum average displacement error} ($\text{minADE}_{K}$) computes the average L2 distance between the ground-truth trajectory and the best predicted trajectory. The \textit{miss rate} (MR) reflects the percentage of scenarios for which the endpoints of none of the predicted trajectories are within \SI{2.0}{\meter} of the ground truth's endpoint. Finally, the main metric for the benchmark, the \textit{brier-minFDE} is defined as
\begin{equation}
    \text{b-minFDE}_{K} = (1-\pi)^2 + \text{minFDE}_{K}, 
\end{equation}
where $\pi$ is the predicted probability for the best predicted trajectory. This metric captures the ability to both predict an accurate future trajectory and assign a high probability to the trajectory that best matches the ground truth, which is relevant in real-world scenarios. The default setting for our experiments is $K=6$, but we also report $K=1$ for test set evaluations, which picks the mode with the highest probability.

\begin{table*}[t]
    \centering
    \renewcommand{\tabcolsep}{2pt}
    \small
    \begin{tabularx}{0.8\linewidth}{s{1.8cm}s{1.6cm}s{1.6cm}YYYY} \toprule
        Decoder-only & Overpredict & Refine & \textbf{b-minFDE\textsubscript{6}}\down & minFDE\textsubscript{6}\down & minADE\textsubscript{6}\down &
        MR\textsubscript{6}\down \\
        \midrule
        \xmark & N/A & N/A & 1.874 & 1.253 & \textbf{0.720} & 0.157 \\
        \midrule
        \cmark & \xmark & \xmark & 1.838 & 1.198 & 0.745 & 0.145 \\
        \cmark & \cmark & \xmark & 1.838 & 1.193 & 0.728 & 0.146 \\
        \cmark & \xmark & \cmark & 1.835 & 1.218 & 0.751 & 0.150 \\
        \cmark & \cmark & \cmark & \textbf{1.807} & \textbf{1.176} & 0.722 & \textbf{0.144} \\
        \bottomrule
    \end{tabularx}
    \caption{\textbf{Ablation study.} We demonstrate the effectiveness of (1) using our \textit{decoder-only} approach instead the encoder-decoder baseline~\cite{zhou2023qcnet}, (2) the \textit{overprediction} objective, and (3) the \textit{refinement} module. Evaluation on the Argoverse2 \textit{val} set~\cite{wilson2021argoverse2}.}
    \label{tab:main_results}
\end{table*}

\PAR{Implementation details.} We implement DONUT within the publicly released code repository of QCNet~\cite{zhou2023qcnet}. In our decoder-only model, we use sub-trajectories of $T_\text{sub}=10$ time steps, \ie, one second. Each token can interact with other agents and map elements that are within $r = \SI{50}{\meter}$ of its reference point. DONUT is trained using the AdamW optimizer~\cite{loshchilov2017adamw}, with a weight decay of $10^{-4}$, an initial learning rate of $5\cdot10^{-4}$, and a cosine learning rate decay schedule. We train for 60 epochs with batches of 8 scenes per GPU across 4 H100 GPUs, and we accumulate gradients to obtain a virtual batch size of 64. We use an embedding dimension of $D=128$ and a dropout rate of 10\%. 

\PAR{Encoder-decoder baseline.} As described in \cref{sec:method:baseline}, we use QCNet~\cite{zhou2023qcnet} as our encoder-decoder baseline. For this model, we use the publicly released code and weights.

\subsection{Main results}
\label{sec:experiments:main_results}
First, we assess the impact of replacing the encoder-decoder approach with our decoder-only setup. The results in \cref{tab:main_results} demonstrate that switching to our decoder-only setup without using \textit{overprediction} or \textit{refinement} boosts the performance across the main b-minFDE\textsubscript{6} metric, the minFDE\textsubscript{6}, and the MR\textsubscript{6}. This demonstrates that using the same unified model structure for both historical and future trajectories allows the model to make more accurate predictions, as hypothesized. On the minADE\textsubscript{6} metric, however, the encoder-decoder baseline performs slightly better. We attribute this to the fact that the baseline's agent encoder generates a single token per time step, giving it a higher granularity.

Adding the refinement module and overprediction yields a strong performance gain. Interestingly, using overprediction or refinement \textit{separately} does not improve performance significantly. When using the refinement module alone, we observe minor training instabilities, which disappear when also applying overprediction. This suggests that overprediction guides DONUT to converge more stably to a better optimum by forcing it to consider longer temporal horizons, thereby unlocking the refiner's potential. At the same time, the benefit of overprediction is larger when using refinement because the refiner has access to other agents’ proposed trajectories and an updated reference point, effectively reducing the overprediction horizon from 20 steps to 10.

\begin{figure}[t]
    \centering
    \begin{tikzpicture}
	\begin{axis}[
		height=5cm,
		width=8cm,
        grid=both,
        ytick={0, 0.2, 0.4, 0.6, 0.8, 1.0, 1.2},
        xlabel={Future time steps},
        ylabel={minFDE\textsubscript{6}\down},
        legend pos=north west,
        legend cell align=left,
        font=\small
	]
		
	\addplot[mark=x, line width=1pt, RWTHRot100] coordinates {
        (10, 0.112)
        (20, 0.264)
        (30, 0.442)
        (40, 0.673)
        (50, 0.951)
        (60, 1.253)
    };
	\addlegendentry{Encoder-decoder}

	\addplot[mark=x, line width=1pt, RWTHBlau100] coordinates {
        (10, 0.126)
        (20, 0.280)
        (30, 0.456)
        (40, 0.672)
        (50, 0.922)
        (60, 1.177)
	};
	\addlegendentry{Decoder-only (ours)}
	\end{axis}
\end{tikzpicture}
    \caption{\textbf{Performance at different prediction horizons.} Compared to the encoder-decoder baseline, our decoder-only approach makes more accurate predictions at longer prediction horizons.}
    \label{fig:horizon}
\end{figure}
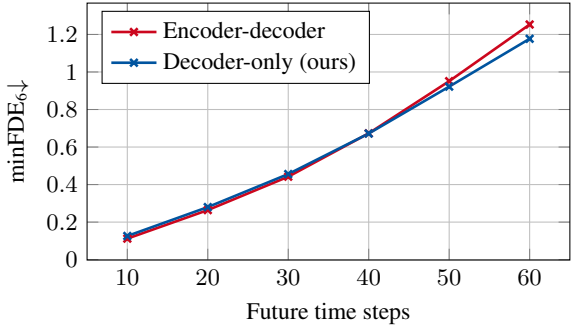

To further assess the impact of the decoder-only model, we examine the final displacement error across various prediction lengths (\cref{fig:horizon}). The results indicate that the decoder-only approach yields significantly improved forecasts for longer prediction intervals. This finding supports our hypotheses that (a) having access to up-to-date previous trajectories in our decoder-only model is particularly beneficial for longer prediction horizons, and (b) overprediction enables the model to better anticipate the far future.

\begin{table*}[t]
    \centering
    \renewcommand{\tabcolsep}{3pt}
    \small
    \begin{tabularx}{\linewidth}{l YYYY YYY} \toprule
        Method & \textbf{b-minFDE\textsubscript{6}}\down & minFDE\textsubscript{6}\down & minADE\textsubscript{6}\down & MR\textsubscript{6}\down & minFDE\textsubscript{1}\down & minADE\textsubscript{1}\down & MR\textsubscript{1}\down 
        \\ 
        \midrule
        THOMAS~\cite{gilles2022thomas}            & 2.16 & 1.51 & 0.88 & 0.20 & 4.71 & 1.95 & 0.64 \\
        Forecast-MAE~\cite{cheng2023forecast-mae} & 2.03 & 1.39 & 0.71 & 0.17 & 4.35 & 1.74 & 0.61 \\
        GoRela~\cite{cui2023gorela}               & 2.01 & 1.48 & 0.76 & 0.22 & 4.62 & 1.82 & 0.66 \\
        HeteroGCN~\cite{gao2023heterogcn}         & 2.00 & 1.37 & 0.73 & 0.18 & 4.53 & 1.79 & 0.59 \\
        MTR~\cite{shi2022mtr}                     & 1.98 & 1.44 & 0.73 & 0.15 & 4.39 & 1.74 & 0.58 \\
        GANet~\cite{wang2023ganet}                & 1.96 & 1.34 & 0.72 & 0.17 & 4.48 & 1.77 & 0.59 \\
        DySeT~\cite{pourkeshavarz2024dyset}       & 1.93 & 1.28 & 0.67 & 0.16 & 4.41 & 1.76 & 0.61 \\
        QCNet~\cite{zhou2023qcnet}                & 1.91 & 1.29 & 0.65 & 0.16 & 4.30 & 1.69 & 0.59 \\
        ProphNet~\cite{wang2023prophnet}          & 1.88 & 1.32 & 0.66 & 0.18 & 4.77 & 1.76 & 0.61 \\
        SmartRefine~\cite{zhou2024smartrefine}    & 1.86 & 1.23 & \underline{0.63} & 0.15 & 4.17 & \underline{1.65} & 0.58 \\
        DeMo~\cite{zhang2024demo}                 & \underline{1.84} & \underline{1.17} & \textbf{0.61} & \textbf{0.13} & \textbf{3.74} & \textbf{1.49} & \underline{0.55} \\
        \midrule
        \grey{MacFormer*~\cite{feng2023macformer}} & \grey{1.91} & \grey{1.38} & \grey{0.70} & \grey{0.19} & \grey{4.69} & \grey{1.84} & \grey{0.61} \\
        \grey{HeteroGCN*~\cite{gao2023heterogcn}}  & \grey{1.90} & \grey{1.34} & \grey{0.69} & \grey{0.18} & \grey{4.40} & \grey{1.72} & \grey{0.59} \\
        \grey{QCNet*~\cite{zhou2023qcnet}}         & \grey{1.78} & \grey{1.19} & \grey{0.62} & \grey{0.14} & \grey{3.96} & \grey{1.56} & \grey{0.55} \\
        \grey{SEPT*~\cite{lan2024sept}}            & \grey{1.74} & \grey{1.15} & \grey{0.61} & \grey{0.14} & \grey{3.70} & \grey{1.48} & \grey{0.54} \\
        \grey{DeMo*~\cite{zhang2024demo}}          & \grey{1.73} & \grey{1.11} & \grey{0.60} & \grey{0.12} & \grey{3.70} & \grey{1.49} & \grey{0.55} \\
        \midrule
        DONUT (ours)                        & \textbf{1.79} & \textbf{1.16} & \underline{0.63} & \underline{0.14} & \underline{4.06} & 1.66 & \textbf{0.54} \\

        \bottomrule
    \end{tabularx}
    \caption{\textbf{Comparison to state of the art.} We compare to published methods on the \textit{test} set of the Argoverse 2 leaderboard~\cite{wilson2021argoverse2}, and demonstrate that DONUT achieves state-of-the-art performance on the main b-minFDE\textsubscript{6} metric. * Denotes the use of model ensembling.}
    \label{tab:sota_comparison}
\end{table*}

\begin{figure*}[t]
    \centering
    \begin{subfigure}{\linewidth}
    \includegraphics[angle=-90,width=0.33\linewidth]{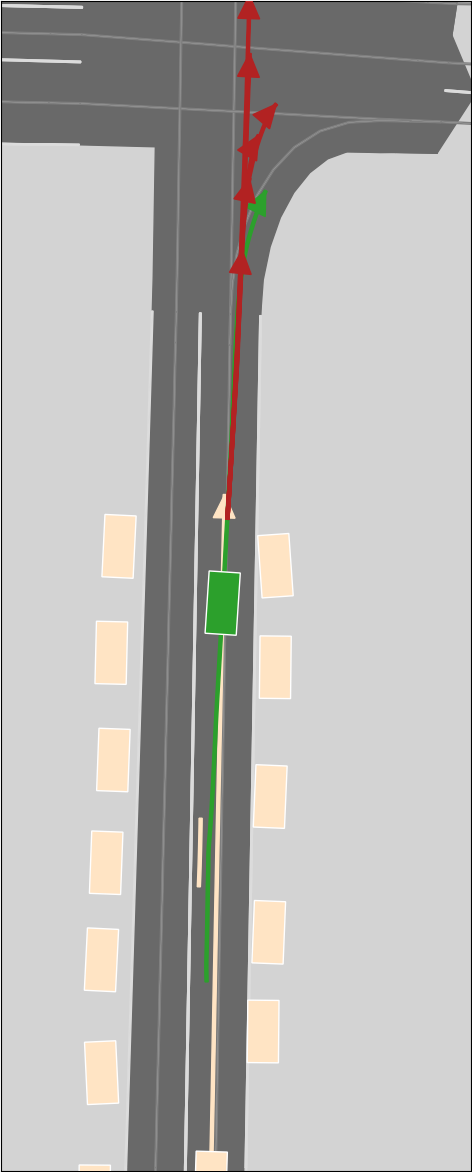}
    \includegraphics[angle=-90,width=0.33\linewidth]{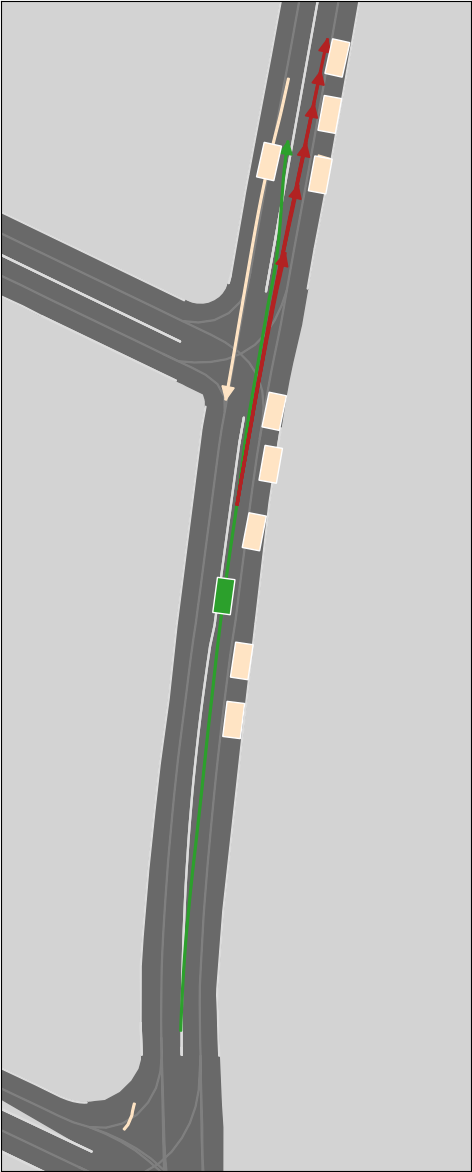}
    \includegraphics[angle=-90,width=0.33\linewidth]{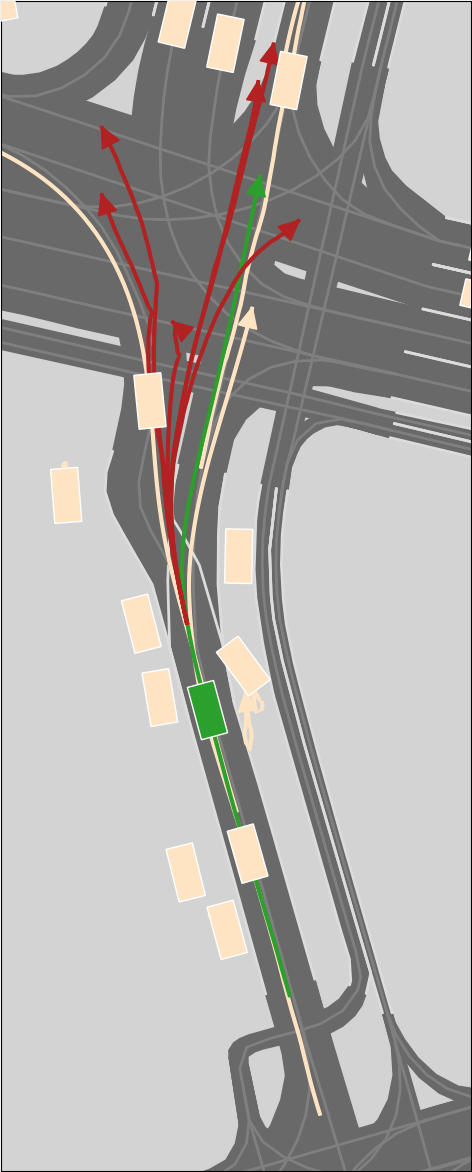}
    \caption{Encoder-decoder}
    \end{subfigure}
    \begin{subfigure}{\linewidth}
    \includegraphics[angle=-90,width=0.33\linewidth]{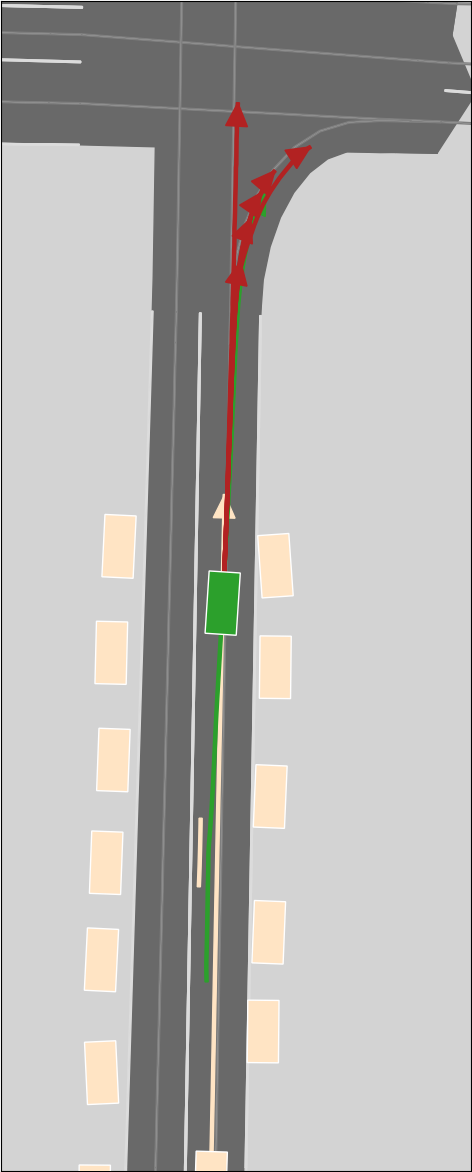}
    \includegraphics[angle=-90,width=0.33\linewidth]{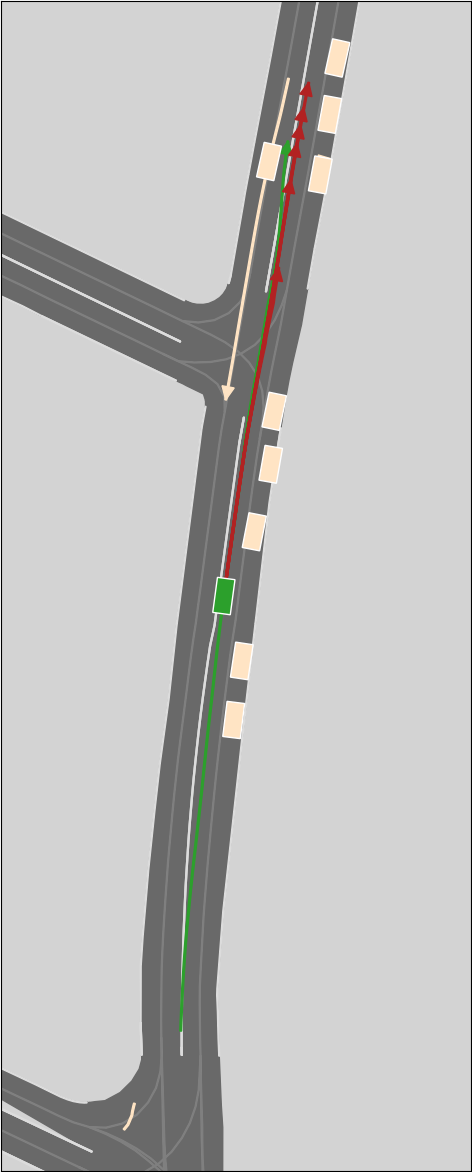}
    \includegraphics[angle=-90,width=0.33\linewidth]{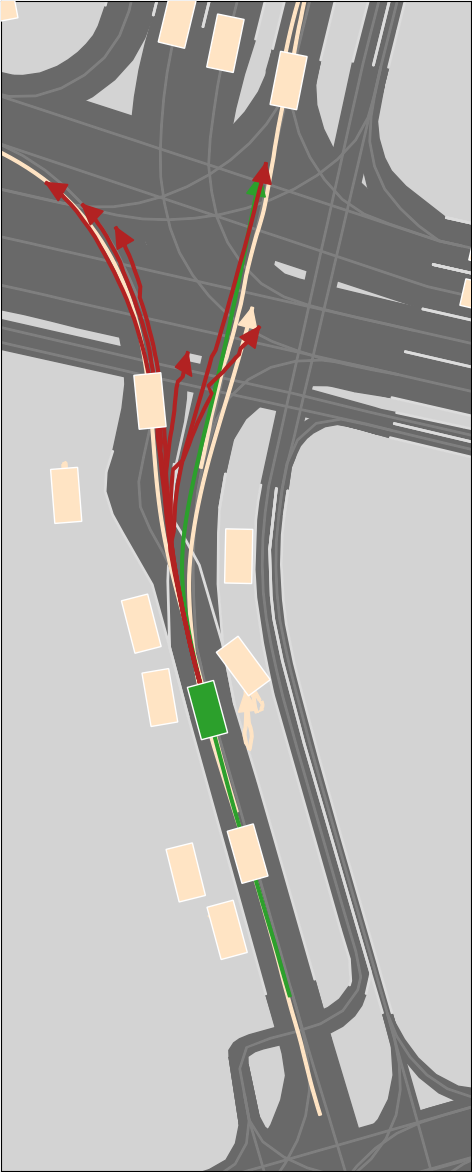}
    \caption{Decoder-only (ours)}
    \end{subfigure}
    \caption{\textbf{Qualitative results.} Green trajectories (\protect\tikz \protect\draw[-{Triangle[scale=1]}, thick, QualitativeGreen] (0,0.1) to[out=0, in=160] (0.5,0);) show the ground-truth historical and future trajectory of the agent of interest, red trajectories (\protect\tikz \protect\draw[-{Triangle[scale=1]}, thick, QualitativeRed] (0,0.1) to[out=0, in=160] (0.5,0);) visualize the $K=6$ future predictions. Boxes show the location of the agents at the start of the predicted trajectory. Overall, we observe that DONUT predicts more accurate trajectories than the encoder-decoder baseline.}
    \label{fig:qualitative}
\end{figure*}

\subsection{Comparison to state of the art}
To assess the effectiveness of DONUT compared to existing models, we submit it to the Argoverse 2 \textit{test} leaderboard and report the results in \cref{tab:sota_comparison}. The results demonstrate that DONUT outperforms all other non-ensemble models on the main metric, the b-minFDE\textsubscript{6}, by considerable margins. As such, it obtains new state-of-the-art results. This demonstrates that DONUT can accurately predict the endpoint of the future trajectory with one of its 6 predictions, and can assign a high probability to the predicted trajectory that is closest to the ground truth. On other metrics, we also achieve new state-of-the-art results or obtain highly competitive scores which shows the strength of our approach. 

DONUT only slightly underperforms DeMo~\cite{zhang2024demo} on the minADE\textsubscript{1} and minFDE\textsubscript{1} metrics, but these unimodal metrics are not suitable to evaluate multimodal predictions. 
An optimal model would always predict the \emph{mean} of the future distribution and assign it the highest probability, which often is very unlikely in reality. 
To clarify, consider an example where a car approaches an intersection where turning left and turning right is equally likely.
Here, the optimal prediction for minADE\textsubscript{1} and minFDE\textsubscript{1} would be to stop in the middle of the intersection.
In contrast, DONUT does achieve state-of-the-art results on the MR\textsubscript{1} metric, which would penalize a mean prediction, as stopping in the middle of the road will most likely not be close to the actual future.

\subsection{Qualitative results}
\label{sec:experiments:qualitative}
We present a few selected qualitative examples contrasting the encoder-decoder baseline with DONUT in \cref{fig:qualitative}. The left example shows the car approaching a normal intersection. While DONUT manages to correctly predict the right turn with most of its modes, QCNet turns too late and heads into oncoming traffic. The middle example visualizes a scenario where the agent has to drive close to the middle of the road as cars are parked on the side. QCNet predicts that the car would go back to the middle of its lane far to early, where it would hit the parked cars, while DONUT manages to keep a safe distance. Both examples showcase the effectiveness of our regular reference point updates during unrolling the future, allowing the model to better take into account scene elements at future time steps.

The third example depicts a particularly complex intersection with many crossing lanes and uncommon polygon shapes. Our decoder-only approach predicts most modes turning left, following the trajectory of the previous car. A single trajectory also drives straight and matches the ground-truth observation really well, while another prediction stops before the intersection and the last one behaves weirdly. In contrast, the encoder-decoder baseline has huge issues with this scene. The two modes going straight leave the street at their endpoint. A third mode turns right in the middle of the intersection and heads into the oncoming traffic. The other three modes turn left, where one overshoots the curve and another shows a strong discontinuity where QCNet splits the prediction into the recurrent parts. Further examples are in the supplemental material.

\section{Conclusion}

In this work, we introduce DONUT, an autoregressive decoder-only model for motion forecasting. By processing historical and future trajectories with a unified model, DONUT makes predictions in a consistent manner and is regularly provided with up-to-date information, unlike existing encoder-decoder methods. Furthermore, by employing an innovative overprediction strategy which guides the model to better anticipate the future, DONUT achieves state-of-the-art results on the Argoverse 2 single-agent motion forecasting benchmark for non-ensemble methods.

Our experiments highlight the advantages of the decoder-only paradigm, especially for complex, longer-term predictions. As such, we expect that this approach will generalize well to more complex forecasting scenarios, and we encourage future work in this direction.

\PAR{Acknowledgments.} This work was partially funded by the BMBF project  6GEM (16KISK036K). The authors gratefully acknowledge the Gauss Centre for Supercomputing e.V.\ (\httpsurl{www.gauss-centre.eu}) for funding this project by providing computing time on the GCS Supercomputer JUWELS at Jülich Supercomputing Centre (JSC). Further computational resources were provided by the German AI Service Center WestAI under project rwth1815 and by RWTH Aachen University under project rwth1825.

\newpage
\clearpage

{
    \small
    \bibliographystyle{ieeenat_fullname}
    \bibliography{main}
}

\newpage

\appendix

\maketitlesupplementary

\section{LineAttention}

In preliminary experiments, we found it beneficial to give the agent better access to road elements. Our baseline, QCNet, uses the beginning of a road polyline as the reference point for relative positional encodings. In addition to this, we add an encoding for relative information between the agent and its closest point on each map polyline, which we call \textit{LineAttention}. However, for the final model, this procedure only had a tiny effect, decreasing minFDE from $1.181$ to $1.176$.

\section{Efficiency Analysis}

We measure inference time and show the number of parameters in \cref{tab:perf}. Due to the temporal unrolling, switching from the baseline to decoder-only almost triples the inference time. The refinement layer roughly doubles the number of successive operations and thus also the inference time. Training times behave similarly. Overprediction has negligible impact on efficiency and is dropped for inference. Note, however, that we did not focus on optimizing the code for efficiency, but instead on improving the prediction accuracy.

\begin{table}[H]
    \centering
    \small
    \renewcommand{\tabcolsep}{2pt}
    \begin{tabularx}{1.0\linewidth}{s{1.1cm}s{1.1cm}YY} \toprule
        DONUT & Ref. & Inference time (ms) & Num.~parameters \\
        \midrule
        \xmark & N/A & 23.7 & 7.7M \\
        \midrule
        \cmark & \xmark & 65.7 & 5.2M \\
        \cmark & \cmark & 129.0 & 9.0M \\
        \bottomrule
    \end{tabularx}
    \caption{\textbf{Efficiency analysis} on an Nvidia RTX 4090 GPU.}
    \label{tab:perf}
\end{table}

\section{Difficult Scenes}

To assess DONUT on more challenging scenarios, we evaluate it on Argoverse 2 trajectories with a ground-truth future turn of at least 45$^\circ$ in \cref{tab:turn}. The relative improvement with respect to the encoder-decoder baseline becomes notably larger than on the full dataset (14.6\% vs.~6.1\% minFDE), showing that DONUT's periodic updates are especially helpful in complex situations.

\begin{table}[H]
    \centering
    \renewcommand{\tabcolsep}{2pt}
    \footnotesize
    \begin{tabularx}{1.0\linewidth}{s{1.1cm}s{0.9cm}s{0.6cm}s{1.4cm}YYY} \toprule
        DONUT & Overp. & Ref. & b-minFDE & minFDE & minADE & MR \\
        \midrule
        \xmark & N/A & N/A & 3.008 & 2.394 & 1.176 & 0.362 \\
        \midrule
        \cmark & \xmark & \xmark & 2.766 & 2.129 & 1.147 & 0.308 \\
        \cmark & \cmark & \xmark & 2.725 & 2.078 & 1.109 & 0.308 \\
        \cmark & \xmark & \cmark & 2.757 & 2.148 & 1.160 & 0.329 \\
        \cmark & \cmark & \cmark & \textbf{2.672} & \textbf{2.043} & \textbf{1.092} & \textbf{0.295} \\
        \bottomrule
    \end{tabularx}
    \caption{\textbf{Results on Argoverse 2, only considering turns $> 45^{\circ}$.}}
    \label{tab:turn}
\end{table}

\section{Tokenizer Details}

\begin{figure}[H]
    \input{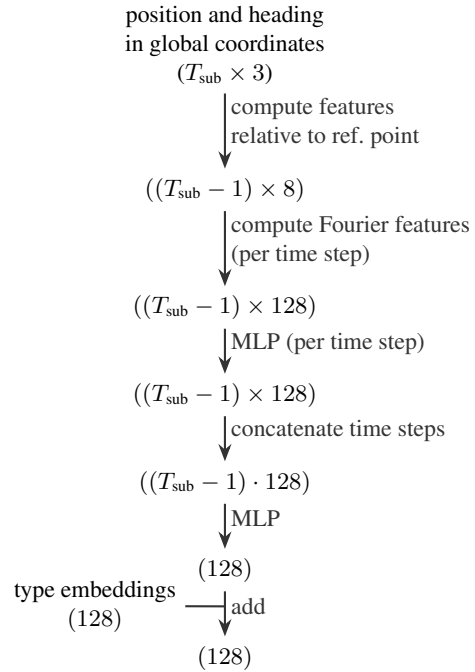}
    \def\shortdist{0.6}
    \def\longdist{1}
    \def\labeldist{0}
    \centering
    \begin{tikzpicture}
        \node[align=center, font=\small] (inp) {position and heading\\in global coordinates\\($T_\text{sub}\times 3$)};
        \node[below=\longdist of inp, align=center, font=\small] (vec) {$((T_\text{sub}-1)\times 8)$};
        \node[below=\longdist of vec, align=center, font=\small] (fourier) {$((T_\text{sub}-1)\times 128)$};
        \node[below=\shortdist of fourier, align=center, font=\small] (mlp1) {$((T_\text{sub}-1)\times 128)$};
        \node[below=\shortdist of mlp1, align=center, font=\small] (cat) {$((T_\text{sub}-1)\cdot 128)$};
        \node[below=\shortdist of cat, align=center, font=\small] (mlp2) {$(128)$};
        \node[below=\shortdist of mlp2, align=center, font=\small] (tok) {$(128)$};
        \node[anchor=east, xshift=-15, yshift=3, at=($(mlp2)!0.5!(tok)$), align=center, font=\small] (type) {type embeddings\\$(128)$};

        \draw [flow] (inp) -- node[label,right,align=left,yshift=3] {compute features\\relative to ref.\ point} (vec);
        \draw [flow] (vec) -- node[label,right,align=left,yshift=3] {compute Fourier features\\(per time step)} (fourier);
        \draw [flow] (fourier) -- node[label,right,yshift=3] {MLP (per time step)} (mlp1);
        \draw [flow] (mlp1) -- node[label,right,yshift=3] {concatenate time steps} (cat);
        \draw [flow] (cat) -- node[label,right,yshift=3] {MLP} (mlp2);
        \draw [flow] (mlp2) -- node[label,right,yshift=3] {add} (tok);
        \draw [black!80, thick] (type) -- ($(mlp2)!0.5!(tok)+(0,0.1)$);
    \end{tikzpicture}
    \caption{\textbf{Detailed tokenizer architecture.}}
    \label{fig:tok}
\end{figure}

In \cref{fig:tok} we visualize our tokenizer's architecture in detail. The $8$-dimensional features for each time step consist of position and heading relative to the reference point, motion vectors, angular motion, velocity, and the difference of the heading and the motion vector direction. Type embeddings describe the object types (\eg, car, bus, pedestrian) present in Argoverse 2.

\section{Failure Cases}

\begin{figure*}
    \centering
    \includegraphics[width=0.19\linewidth]{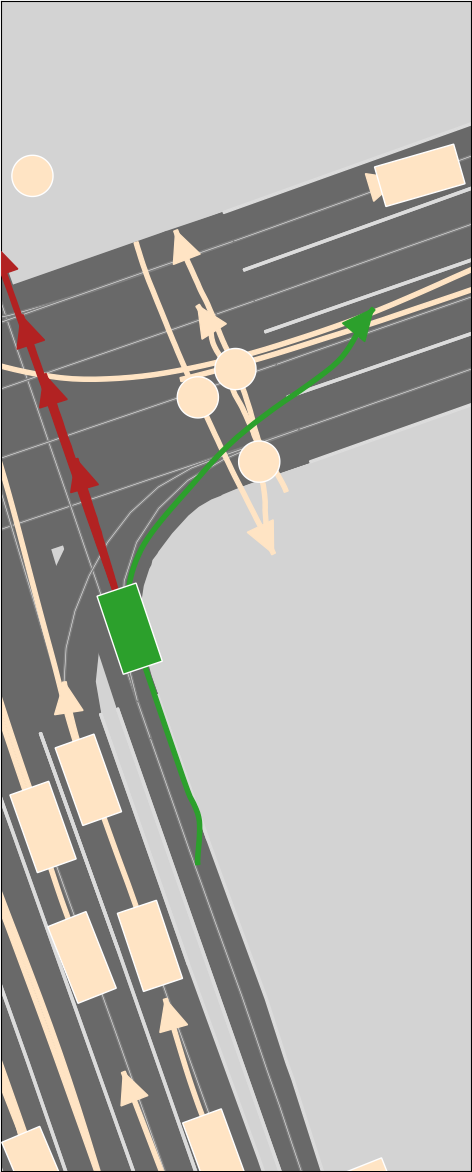}
    \includegraphics[width=0.19\linewidth]{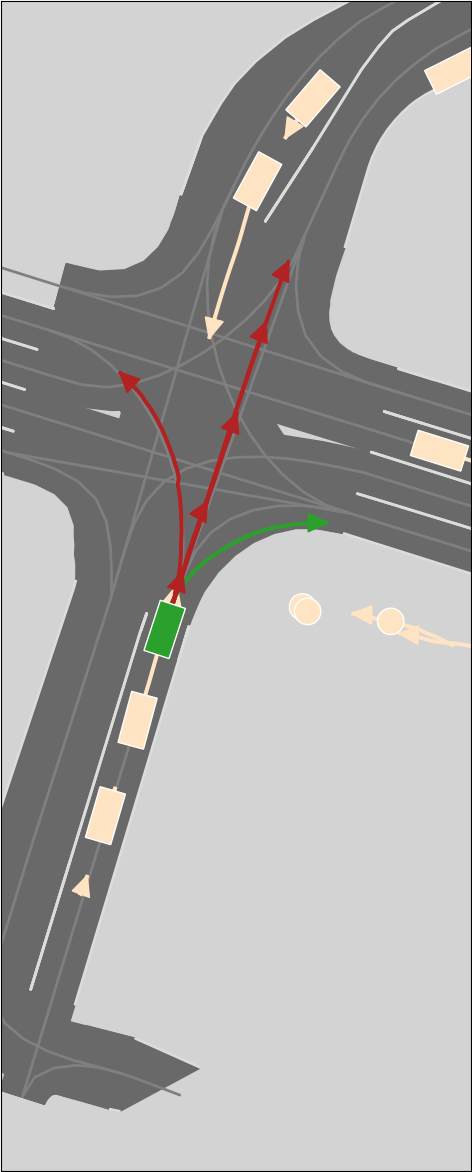}
    \includegraphics[width=0.19\linewidth]{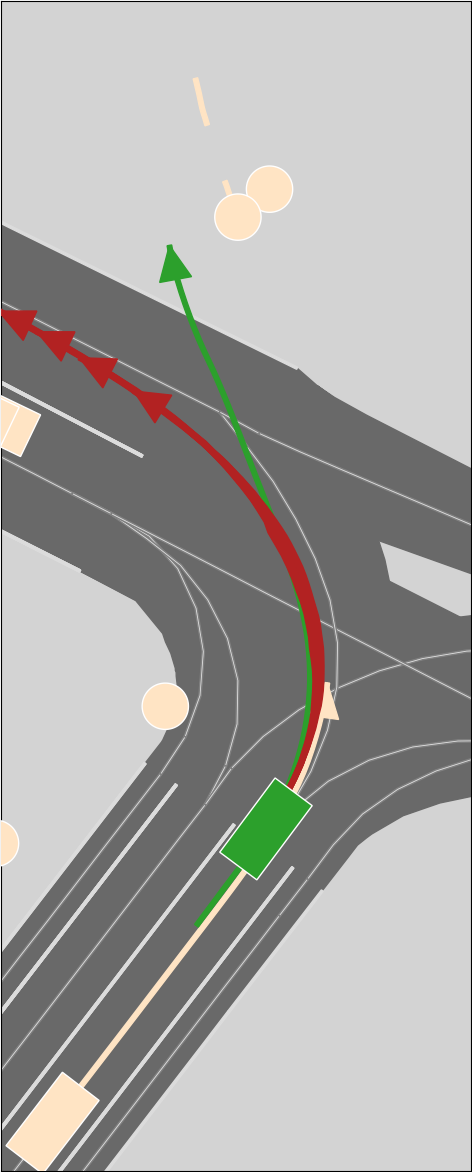}
    \includegraphics[width=0.19\linewidth]{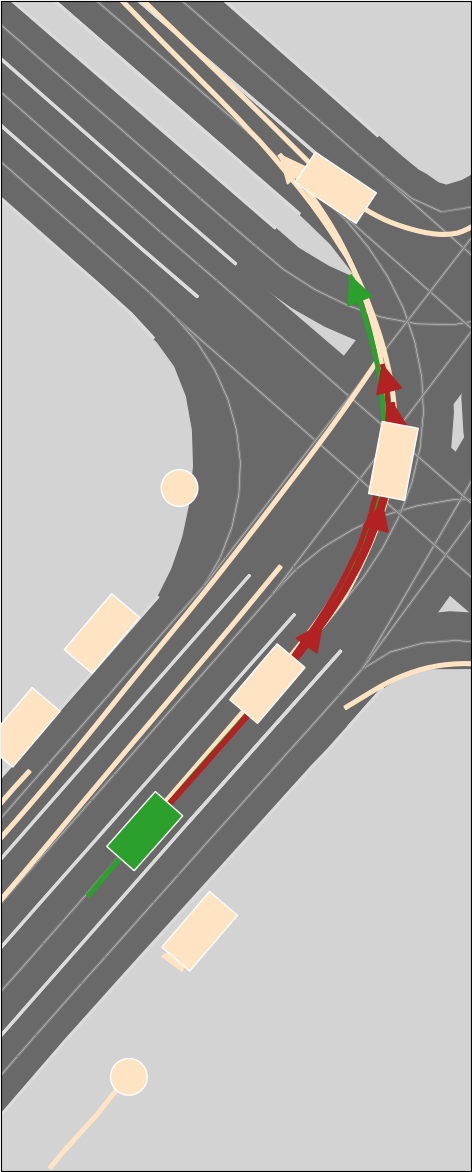}
    \includegraphics[width=0.19\linewidth]{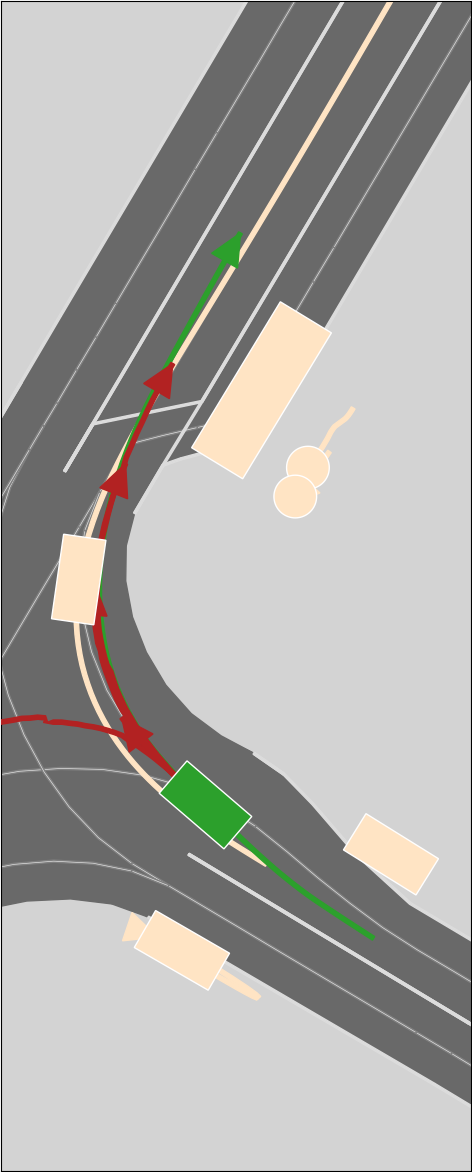}
    \includegraphics[width=0.19\linewidth]{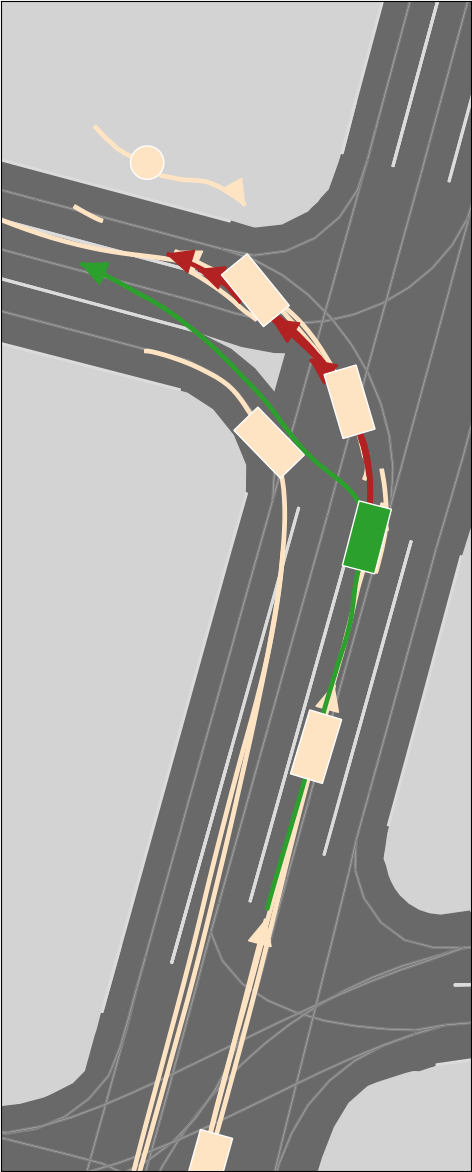}
    \includegraphics[width=0.19\linewidth]{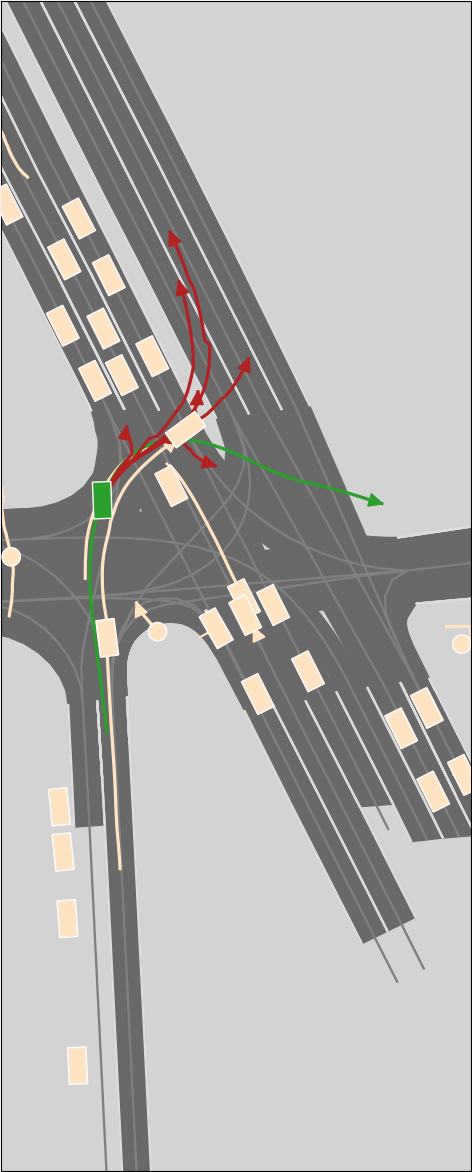}
    \includegraphics[width=0.19\linewidth]{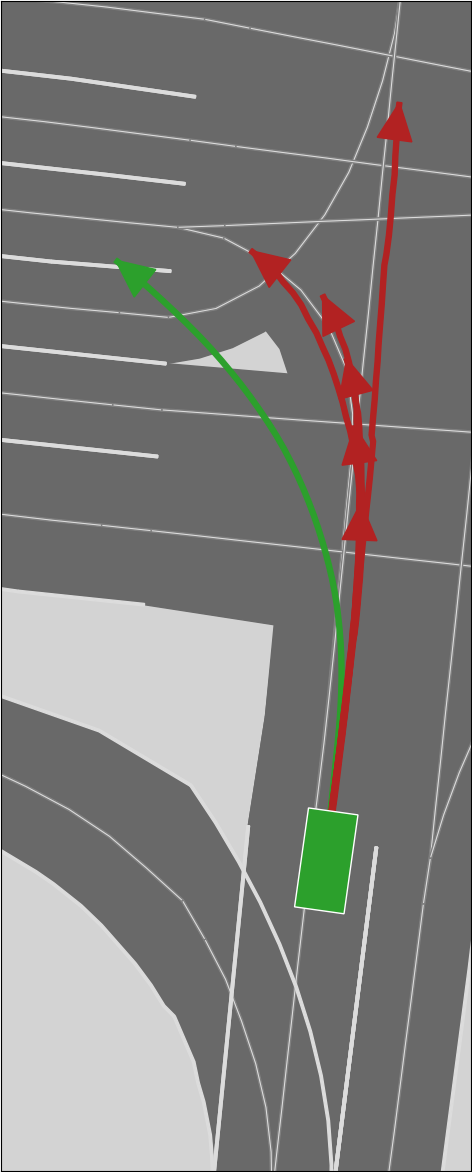}
    \includegraphics[width=0.19\linewidth]{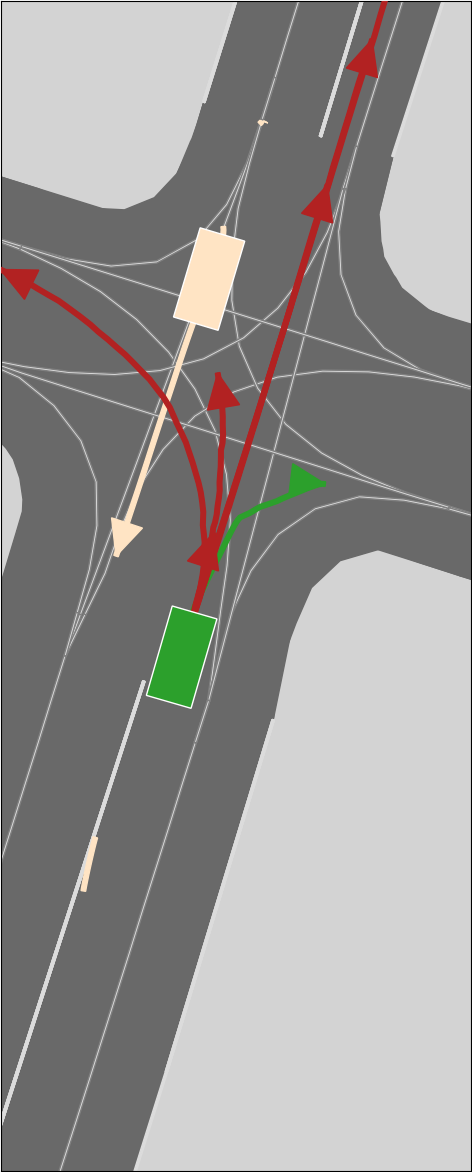}
    \includegraphics[width=0.19\linewidth]{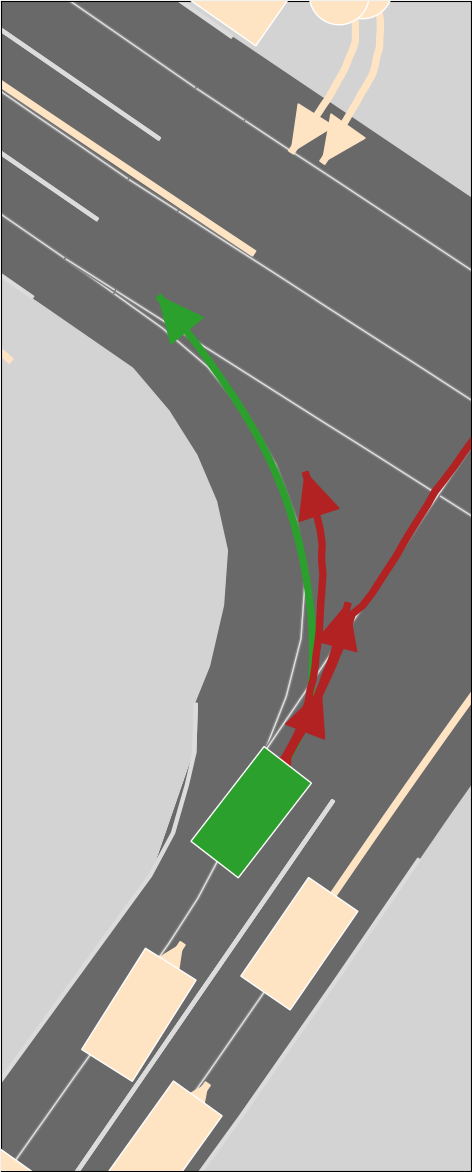}
    \caption{\textbf{Failure cases of DONUT.}}
    \label{fig:fail}
\end{figure*}

We manually examined 100 scenes with a minFDE $>$ \SI{5}{m}. Most errors are caused by predictions being too slow (27\%) or too fast (19\%), or missing a turn (19\%). Additionally, 27\% had rare ground-truth events, \eg, vehicles moving off the road or maneuvering illegally. We visualize a few scenes in \cref{fig:fail}.
\section{Additional Qualitative Results}

We provide additional non-cherrypicked qualitative results in Figs.~\ref{fig:supp:0} to \ref{fig:supp:15}.

\foreach \i in {0,...,15}{

    \begin{figure*}[b]
        \centering
        \begin{subfigure}{0.19\linewidth}
            \centering
            \includegraphics[width=\linewidth]{img/additional/\i_qcnet.png}
            \caption{Encoder-decoder}
        \end{subfigure}
        \begin{subfigure}{0.19\linewidth}
            \centering
            \includegraphics[width=\linewidth]{img/additional/\i_donut_xx.png}
            \caption{DONUT Overp.~\xmark{} Ref.~\xmark}
        \end{subfigure}
        \begin{subfigure}{0.19\linewidth}
            \centering
            \includegraphics[width=\linewidth]{img/additional/\i_donut_ox.png}
            \caption{DONUT Overp.~\cmark{} Ref.~\xmark}
        \end{subfigure}
        \begin{subfigure}{0.19\linewidth}
            \centering
            \includegraphics[width=\linewidth]{img/additional/\i_donut_xr.png}
            \caption{DONUT Overp.~\xmark{} Ref.~\cmark}
        \end{subfigure}
        \begin{subfigure}{0.19\linewidth}
            \centering
            \includegraphics[width=\linewidth]{img/additional/\i_donut_or.png}
            \caption{DONUT Overp.~\cmark{} Ref.~\cmark}
        \end{subfigure}
        \caption{\textbf{Additional qualitative results.}}
        \label{fig:supp:\i}
    \end{figure*}
}

\end{document}